\documentclass[runningheads]{llncs}
\usepackage[T1]{fontenc}
\usepackage[dvipsnames]{xcolor}
\usepackage{makecell}
\usepackage{graphicx,verbatim}
\usepackage{diagbox}
\usepackage{tikz}
\usetikzlibrary{positioning, fit, arrows, shapes.geometric}
\usepackage{standalone}

\tikzset{
    state/.style={circle, draw=black, fill=blue!10, thick, minimum size=1.cm, node distance=1.cm},
    arrow/.style={-stealth, thick},
    bend arrow/.style={-stealth, thick, bend angle=60, bend left},
    state1/.style={circle, draw=black, fill=green!10, thick, minimum size=1.cm, node distance=1.cm},
    rec/.style={rectangle, draw=black}
}
\usepackage{hyperref}
\hypersetup{
    colorlinks=true,
    linkcolor=blue,
    citecolor=blue,
    urlcolor=blue,
    }
\usepackage{xcolor}
\usepackage{bm}
\usepackage{amsmath,amssymb}

\begin{document}
\title{Unsupervised anomaly detection using Bayesian flow networks: application to brain FDG PET in the context of Alzheimer's disease }
\titlerunning{Unsupervised anomaly detection using Bayesian flow networks}
%

\author{Hugues Roy\inst{1}, Reuben Dorent\inst{1,2}, \and Ninon Burgos\inst{1}}  
\index{Roy, Hugues}
\index{Dorent, Reuben}
\index{Burgos, Ninon}
\authorrunning{H. Roy et al.}
\institute{
Sorbonne Université, Institut du Cerveau - Paris Brain Institute - ICM, CNRS, Inria, Inserm, AP-HP, Hôpital de la Pitié Salpêtrière, F-75013, Paris, France \\ \and
MIND Team, Inria Saclay, Université Paris-Saclay, Palaiseau, France \\
}
    
\maketitle             
\begin{abstract}

Unsupervised anomaly detection (UAD) plays a crucial role in neuroimaging for identifying deviations from healthy subject data and thus facilitating the diagnosis of neurological disorders. In this work, we focus on Bayesian flow networks (BFNs), a novel class of generative models, which have not yet been applied to medical imaging or anomaly detection. BFNs combine the strength of diffusion frameworks and Bayesian inference. We introduce AnoBFN, an extension of BFNs for UAD, designed to: i) perform conditional image generation under high levels of spatially correlated noise, and ii)  preserve subject specificity by incorporating a recursive feedback from the input image throughout the generative process.
We evaluate AnoBFN on the challenging task of Alzheimer's disease-related anomaly detection in FDG PET images. 
Our approach outperforms other state-of-the-art methods based on VAEs ($\beta$-VAE), GANs (f-AnoGAN), and diffusion models (AnoDDPM), demonstrating its effectiveness at detecting anomalies while reducing false positive rates.

\keywords{Generative models \and Bayesian Flow Networks \and Anomaly detection \and Neuroimaging} 

\end{abstract}

\section{Introduction}

Pixel-wise anomaly detection is a key problem in medical image computing. The current state-of-the-art approach is supervised learning (segmentation), which requires large and manually annotated datasets. Alternatively, unsupervised anomaly detection (UAD) has raised significant interest as it bypasses the need for annotation by training a generative model on healthy data  \cite{Cai2025MedIAnomalyComparative}. 
At inference, the model reconstructs a pseudo-healthy version of a given image, and anomalies are identified by comparing the original and reconstructed images, enabling the detection of lesions without prior knowledge of their specific appearance. 

Classical UAD generative models include the f-AnoGAN \cite{schlegl_f-anogan_2019} and variational autoencoders (VAEs) \cite{kingma_auto-encoding_2022,Baur2021AutoencodersUnsupervised,Hassanaly2024PseudohealthyImage}.
However, these models, trained only on healthy data, are sensitive to distribution shifts and often produce unreliable encodings for anomalous inputs \cite{cai2024rethinking}. Consequently, normal regions may be misreconstructed, leading to a loss of subject specificity and a large rate of false positives. 

Denoising diffusion models \cite{ho_denoising_2020,song_score-based_2021,kingma_variational_2021} address the encoding limitation by using a latent representation based on noisy versions of the data. 
These models have demonstrated competitive performance in chest x-ray \cite{wolleb_diffusion_2022} or brain MRI \cite{pinaya_fast_2022,wolleb_diffusion_2022,wyatt_anoddpm_2022,yu_adversarial_2023} applications. 
A key limitation of diffusion models is their tendency to change regions even when no anomalies are present. Higher noise levels are required to eliminate larger anomalies but reduce subject's specificity. 
Therefore, an optimal balance must be achieved between effective anomaly removal and preservation of subject's specificity.

In this context, we leverage the use of Bayesian flow networks (BFNs)~\cite{graves_bayesian_2024} , a novel class of generative models based on diffusion models, introducing a Bayesian latent representation on the parameters of a data distribution rather than on the data directly. The latent parameter variable allows information aggregation of past generated samples and uses Bayesian inference to update the current state.
BFNs have shown promise in 3D molecule modeling  \cite{atkinson_protein_2024}, material generation \cite{wu2025periodicbayesianflowmaterial}, and crystal generation \cite{ruple2025symmetryawarebayesianflownetworks} but, to our knowledge, have not yet been used in medical imaging or for anomaly detection.

We present AnoBFN, the first use of BFNs for unsupervised anomaly detection. First, we adapt BFNs to anomaly detection, introducing a novel probabilistic framework for this task. Second, we use simplex noise with a new accuracy schedule for conditional generation, preserving subject specificity under high spatially correlated noise. Third, we introduce a recursive Bayesian update scheme that fuses the original image with the model's prediction to retain input information. Finally, we validate our approach on FDG-PET scans for Alzheimer’s disease detection, a clinically relevant task characterized by diffuse metabolic anomalies and absence of ground truth. To enable quantitative evaluation, we generate synthetic anomalies that replicate disease-like abnormalities and compare our method against state-of-the-art unsupervised 
anomaly detection models.

\section{Bayesian flow networks}

\begin{figure}[tp]
  \centering
\includegraphics[width=\textwidth]{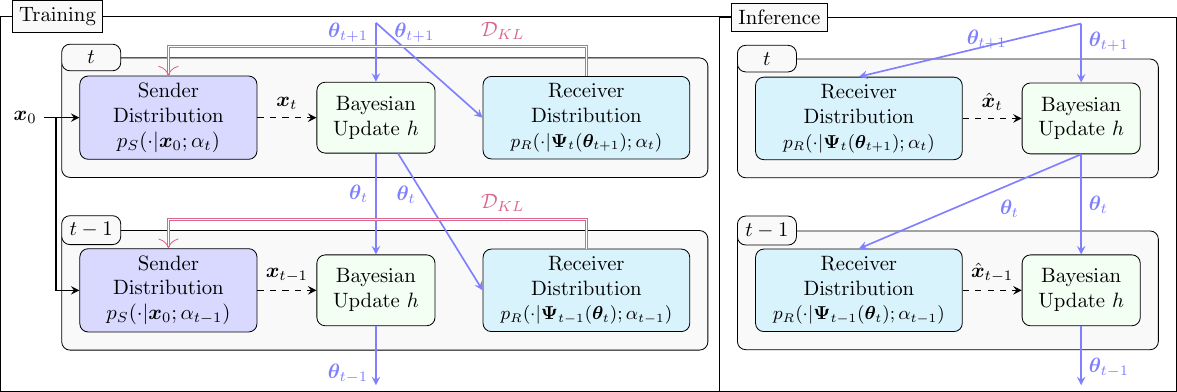}
  \caption{Training and inference phases of the original Bayesian flow networks. Dashed lines correspond to samples from distributions, notations are defined in the main text.}
  \label{fig:bfn_principle}
\end{figure}

Bayesian flow networks (BFNs)~\cite{graves_bayesian_2024} extend denoising diffusion models (DDMs)~\cite{ho_denoising_2020} by operating on the parameters of data distributions rather than directly on noisy data samples. BFNs aim to learn the parameters $\vtheta_0$ of an \textit{input distribution} that approximates the true data distribution $p_{\text{data}}$. This is done by an iterative process of steps $t = T, \dots, 0$ that refines the parameters $ \vtheta_t = \{\vmu_t, \rho_t \} $, considered as latent variables, of the input distribution, where $p_I(\vx_{t}) = \mathcal{N}(\vx_{t};\vmu_t, \rho_t^{-1} \bm{I})$ and $\vx_t$ denotes the noisy data at time $t$. To update $ \vtheta_t $, the authors propose a principled formulation based on Bayes' rule, leveraging observed $ \vx_{t}  $ or estimated $ \hat{\vx}_{t} $ data. The original BFN framework is illustrated in Fig.~\ref{fig:bfn_principle}.
During training, the observed data $ \vx_{t} $ is drawn from a \textit{sender distribution }$ p_S(\vx_t|\vx_0,\alpha_t)=\mathcal{N}(\vx_t;\vx_0,\alpha_t^{-1} \bm{I}) $ by adding noise to the true data $ \vx_0 $ according to a predefined accuracy parameter $\alpha_t = \frac{d \beta(t)}{dt}$ derived from the \textit{accuracy schedule} $\beta_t$, as in DDMs. In the original framework, the noise level initially with $t$, then decreases, reaching a minimum at $t=0$, where the distribution closely approximates true data. This design enables sampling from $p_{\text{data}}$ by initializing the generation process from a zero-mean prior. Then, Bayesian inference is used to update the parameters $\vtheta_t$ by sampling from the following \textit{update distribution} $p_U$
\begin{align}
\label{eq:update_distribution1}
    p_U(\vtheta_t \mid \vtheta_{t+1}; \vx_{0}, \alpha_t) & = \mathbb{E}_{p_S(\vx_{t}  \mid \vx_{0}; \alpha_t)} \left[\delta\left(\vtheta_t - h(\vtheta_{t+1}, \vx_{t}, \alpha_t)\right) \right] \enspace ,
\end{align}
where $\delta$ is the Dirac delta function and $h$ the Bayesian update function computing the posterior using Bayes' rule and a Gaussian conjugate prior $\vtheta_{t+1}$
\begin{align}
    h : \vtheta_{t+1}, \vx_t, \alpha_t \mapsto \left(\vmu_t = \frac{\rho_{t+1} \vmu_{t+1} + \alpha_t \vx_t}{\rho_{t+1} + \alpha_t}, \ \rho_t = \rho_{t+1} + \alpha_t \right)  \enspace.
\end{align}
$\vtheta_t$ represents the updated belief about the data after observing $\vx_{t}$. This process mimics Bayesian inference, gradually refining estimates as evidence is gathered. At inference time, the estimated data $ \hat{\vx}_{t} $ is sampled from a \textit{receiver distribution}, which follows the same noise schedule $ \alpha_t $. Specifically, a neural network $ \Psi $ predicts the mean of the receiver distribution using the current parameters $ \vtheta_{t+1} $, which are either sampled from the prior distribution $p_{P}$ (if $ t +1= T $) or the update Bayesian distribution $p_U$. The receiver distribution is defined as  
\begin{equation}
\label{eq:receiver_distribution}
    p_R(\hat{\vx}_{t} \mid \vtheta_{t+1}; \alpha_{t}) = \mathcal{N}( \hat{\vx}_{t};\Psi(\vtheta_{t+1}), \alpha_{t}^{-1} \bm{I}) \enspace.
\end{equation}  
Similarly to Eq.~\eqref{eq:update_distribution1}, the parameters $ \vtheta_{t}$ used at inference  are sampled from the update distribution $p_U$, determined by the previous parameters $\vtheta_{t+1}$, the estimated data $\hat{\vx}_{t}$ sampled from the receiver distribution $ p_R $ defined in Eq.~\eqref{eq:receiver_distribution} and the Bayesian update $h$
\begin{align}
\label{eq:update_distribution}
    p_U(\vtheta_t \mid \vtheta_{t+1};  \alpha_t) & = \mathbb{E}_{p_R(\hat{\vx}_{t} \mid \vtheta_{t+1} ; \alpha_t)} \delta\left(\vtheta_t - h(\vtheta_{t+1}, \hat{\vx}_{t}, \alpha_t)\right)  \enspace .
\end{align}

To ensure that the estimated data $\hat{\vx}_{t}$ from the receiver distribution are faithful, the Kullback-Leibler divergence ($D_{K L}$) between the sender  $p_S$ and receiver $p_R$ distributions is minimized during training across different timesteps and data
\begin{align}
\label{eq:loss_n}
    \mathcal{L}(\Psi) = \mathbb{E}_{\vx_0 \sim p_{\text{data}}} \mathbb{E}_{ p_U(\vtheta_t|\vtheta_T,\vx_0,\beta_{t+1})}  
    D_{K L} \left[ p_S(\cdot|\vx_0,\alpha_t) \big\| p_R(\cdot \mid \vtheta_{t+1}; \alpha_{t})\right]  .
\end{align}

While BFNs have originally been introduced for generation purposes \cite{graves_bayesian_2024}, they have not been adapted to anomaly detection yet.

\section{Methods} 

In this work, we propose AnoBFN, an approach using Bayesian flow networks for unsupervised anomaly detection in images.
Following the standard UAD framework, we first train a generative model to approximate the distribution of healthy data $p_H$, with samples $\vx_{0,H}$. Then, we leverage this trained model to generate healthy versions of images $\vx_{0,A}$ drawn from an abnormal distribution $p_A$, enabling anomaly detection. To achieve this objective, we extend the original BFN method to perform conditional image generation in the presence of anomalies.

Specifically, our contributions focus on two key objectives: [C1] enabling conditional generation under high levels of spatially correlated noise, which are needed to encourage overlapping prior distributions between abnormal scans and their pseudo-healthy counterparts; [C2] preserving subject specificity by incorporating recursive feedback from the input throughout the generative process.

For our first contribution [C1], we \textbf{jointly use simplex noise} and introduce a \textbf{new accuracy schedule}.

\textit{[C1.1] Structured noise.} Gaussian noise is independent and identically distributed, lacking spatial coherence with a flat power spectrum in the Fourier decomposition space and high-frequency spatial variations \cite{wyatt_anoddpm_2022}. In contrast, simplex noise provides gradient-based noise with spatial continuity and structured perturbations. Inspired by \cite{wyatt_anoddpm_2022}, we use structured noise to encourage the denoiser to handle spatially correlated perturbations rather than independent pixel noise.

\begin{figure}[t]
    \centering
    \includegraphics[width=\textwidth, clip, trim={0, 5, 4, 2}]{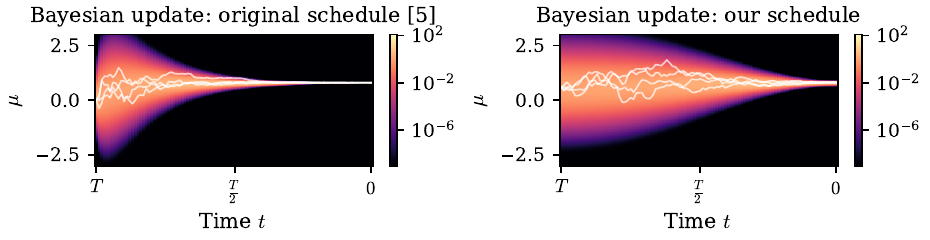}
    \caption{The white curves illustrate sample trajectories of $\mu$, the mean parameters of the latent variable $\vtheta$, under both the schedule from \cite{graves_bayesian_2024} and our accuracy-based schedule. The background shows the probability flow distribution.}
    \label{fig:comp_var_input}
\end{figure}

\textit{[C1.2] Accuracy schedule.} To reconstruct a pseudo-healthy version $\vx_{0,H}$ from an abnormal input $\vx_{0,A}$, the generative process should (i) be conditional and (ii) operate under a high-noise regime to ensure overlap between the prior distributions of healthy and abnormal data. In practice, the mean parameters $\vmu_t$ follow stochastic trajectories defined by the \textit{probability flow distribution} $p_F$ with mean and variance characterized by the accuracy scheldule $\beta(t)$:
\begin{align}
\label{eq:probability_flow}
    p_F(\mu_t | \vx_0; t)  & = p_U(\mu_t | \vtheta_T, \vx_0, \beta(t)) \\
    & = \mathcal{N} \left( \vmu_t \,;\, \frac{\beta(t) \vx_0 + \rho_T \vmu_T}{\rho_T + \beta(t)}, \frac{\beta(t)}{(\rho_T+\beta(t))^2} \bm{I} \right) \enspace .
\end{align}  
Since the original BFN is designed for unconditional generation, the prior distribution of $\vmu_T$  does not exploit subject-specific information, i.e. $\beta(T) = 0$ (Fig.~\ref{fig:comp_var_input}).  In contrast, our goal is to design a scheduler $\beta$ such that: i) the prior distribution depends on the input, i.e. $\beta(T) > 0$; ii) the abnormal and normal prior distributions overlap, i.e. the variance of the prior of $\vmu_T$ is large; and iii) it is well-defined for Bayesian update. Under the prior $\vtheta_T =\{\vmu_T, \rho_T\} = \{\bm{0},1\}$, the condition iii) can be satisfied if there is a function $f \in C([0,T])$ such that $f([0,T]) \subset (0,\frac{1}{4}]$ and for all $ t \in [0,T], \quad f(t)=\frac{\beta(t)}{(1+\beta(t))^2} $. Inspired by \cite{nichol_improved_2021}, a suitable choice shown in Fig.~\ref{fig:comp_var_input} with an initial high variance decreasing over time can be expressed as:
\begin{equation}
\label{eq:cosine_function}
    f(t) = \frac{1}{4} \cos^4 \left( \frac{(T-t)+s}{T(1+s)} \frac{\pi}{2} \right)
\end{equation}  
with $s=0.01$. This introduced scheduler i) preserves information about the input $\vx_0$ at $t = T$, as the initial mean of the probability flow distribution $p_{F}$ is approximately $\frac{ \vx_0}{2}$ and ii) has maximal variance at $t=T$.

Overall, the combination of simplex noise and the modified probability flow enables conditional generation under high spatially correlated noise, encouraging overlapping prior distributions of $\vmu_T$ for normal and abnormal scans.
\

\textit{[C2] Bayesian Update.} Our second key objective is to preserve the subject specificity throughout the generative process at inference time. In classical anomaly detection frameworks, the abnormal image is only used once as input and to compute the residual, i.e., the difference between the input and reconstruction. 
In this work, we leverage the Bayesian update of BFNs to leverage the input image throughout the generation process. 
Specifically, we propose a novel Bayesian update that incorporates both the estimated noisy sample $\hat{\vx}_{t}$ drawn from the learned receiver distribution $ p_R(\hat{\vx}_{t} \mid \vtheta_{t+1}; \alpha_{t}) = \mathcal{N}( \hat{\vx}_{t};\Psi(\vtheta_{t+1}), \alpha_{t}^{-1} \bm{I})$ and the input image $\vx_{0}$, allowing the latent variable to retain information from both sources. 
Therefore, the update distribution also depends on $\vx_0$
\begin{align}
    p_U(\vtheta_{t}| \vtheta_{t+1},\hat{\vx}_{t},\vx_{0}) \enspace .
\end{align}
Similarly to the original framework, the Bayesian update employs a matrix $\valpha_t$ to weight the contribution of $\hat{\vx}_t$. 
To additionally weight the impact of the observed input image $\vx_0$ in the generative process during inference, we introduce an auxiliary weighting parameter $\valpha_{t,A}$ that modulates its contribution.  The Bayesian update function is then defined  as $h : \vtheta_{t+1}, \vx_t, \alpha_t, \vx_{0}, \valpha_{t,A} \mapsto (\vmu_{t},\vrho_{t})$ such that:
\
\begin{equation}  
    \left\{
    \begin{array}{ll}
    \vmu_{t}  & = \frac{1}{\vrho_{t}}( \vrho_{t+1} \odot\vmu_{t+1}  + \valpha_{t} \odot \hat{\vx}_{t} + \valpha_{t,A} \odot \vx_{0}) \\
    \vrho_{t} & = \vrho_{t+1} + \valpha_{t} + \valpha_{t,A} 

    \end{array}
\right.
\end{equation}

Since the parameter $\valpha_{t,A}$ weights the contribution of the input image, $\valpha_{t,A}$ is expected to be larger in normal regions than in abnormal ones. 
To quantify in an unsupervised manner regions that are likely to be abnormal during the inference process, we postulate that abnormal regions are the most varying regions in the generation process. For this reason, we modulate the parameter $\valpha_{t,A}$ relying on the predefined accuracy parameter $\valpha_{t}$ using a scaling metric based on the squared error between the generated mean of the receiver distribution and the input scan, and a time scaling factor ensuring the decay of $\valpha_{t,A}$.
The higher the squared error, the less the update accounts for the input.
For the time scaling, we decided to use the logistic function, with a logistic growth rate $k=30$ and $t_c=0.5$, such that the value of the function's midpoint is at the middle of the generative process
\begin{equation}
    \valpha_{t,A} = \valpha_{t} \odot e^{- (\Psi(\vtheta_{t+1})-\vx_{0})^2} \left(\frac{1}{1+e^{-k(\frac{t}{T}-t_c)}}\right) \enspace .
\end{equation}  

This formulation ensures that the Bayesian update accounts for structural differences between normal and anomalous regions while leveraging prior knowledge from the generative process.

\section{Experiments and results}
In this section, we demonstrate the effectiveness of our approach in detecting anomalies in FDG PET scans associated with Alzheimer's disease (AD). 
A key challenge when evaluating UAD in such context is the absence of ground truth masks of the anomalies, contrary to applications that aim to detect lesions on structural MRI \cite{Baur2021AutoencodersUnsupervised}.
To address this, we assess our method using abnormal data obtained by simulating realistic hypometabolism characteristic of AD \cite{Hassanaly2024EvaluationPseudohealthy}.

\textit{Dataset.} The FDG PET scans used in this study were obtained from the Alzheimer's Disease Neuroimaging Initiative (ADNI) database \cite{weiner_alzheimers_2010,weiner_alzheimers_2017}. We downloaded the scans which frames were co-registered, averaged, and standardized to a uniform resolution. Subsequent image processing was conducted using Clinica~\cite{routier_clinica_2021}, which included linear registration to the MNI space, intensity normalization using the mean PET uptake within the cerebellum and pons, and cropping. The images were further resampled to a grid size of 128×128×128 voxels and rescaled to the range $\left[-1,1\right]$. From each volume, 20 axial slices centered around the central slice were extracted. 
We selected 733 scans from 301 cognitively normal (CN) subjects. First, 80 subjects/scans were isolated as test set. A synthetic image mimicking 30\% AD-induced hypometabolism was generated for each test scan using the method described in \cite{Hassanaly2024EvaluationPseudohealthy}. The healthy images, denoted by test\textsubscript{CN}, were used to evaluate reconstruction performance while their synthetic abnormal counterparts, denoted by test\textsubscript{sAD}, were used to assess anomaly detection performance. 
The remaining scans were split into training and validation sets at the subject level using ClinicaDL \cite{thibeau-sutre_clinicadl_2022}, with stratification based on age and sex, resulting in a training and validation set of 540 and 57 scans, respectively.

\textit{Baselines, implementation details and evaluation metrics.}
We compare our results with the $\beta$-VAE \cite{higgins_beta-vae_2017}, the f-AnoGAN \cite{schlegl_f-anogan_2019}, and the AnoDPPM \cite{wyatt_anoddpm_2022}, so each generative model class is represented in our experiments. 
We used the same Unet~\cite{ronneberger_u-net_2015} architecture from \cite{ho_denoising_2020} for AnoDDPM and AnoBFN.
The structure was modified resulting in an encoder (decoder) composed of 3 downsampling (upsampling) stages with layer width $[C,C,2C]$ with $C=128$. The attention layers are composed of $4$ heads with dimensions $16$.
The optimizer was AdamW \cite{loshchilov_decoupled_2019} with learning rate $1e-4$, weight decay $0.01$ and $(\beta_1, \beta_2) = (0.9, 0.98)$. An exponential moving average of model parameters with a decay rate of $0.9999$ was used for evaluation and sample generation for AnoDDPM and AnoBFN. The total number of learnable parameters was $\sim10$M. The batch size was $30$ and we used gradient clipping to $1$ during the training steps.
Regarding the $\beta$-VAE, we use $\beta=10$, a latent dimension of size $128$, and a symmetrical encoder-decoder architecture using batch normalization and LeakyRelu activation functions with channel multiplication [1,2,6,8]. For the f-AnoGAN, we used the same architecture as the $\beta$-VAE. Code will be available upon manuscript acceptance.

We evaluate reconstruction quality using the mean square error (MSE), peak-to-signal noise ratio (PSNR), and structural similarity (SSIM) computed for healthy scans of the test set test\textsubscript{CN}.
We assess anomaly detection performance using simulated abnormal scans from test\textsubscript{sAD} by computing the intersection over union (IoU) for a threshold of 0.05 corresponding to the mean square synthetic mask's intensity decrease,  and the average precision (AP) at the pixel-level \cite{Cai2025MedIAnomalyComparative}.

\begin{table}[t!]
\caption{Pseudo-healthy reconstruction and anomaly detection performance metrics. Standard deviations are calculated at the subject level. Statistical significance between AnoBFN and alternative methods was determined using a Bonferroni-corrected paired Wilcoxon signed-rank test, $*$ indicating statistically significant differences (p \textless 0.01).}
\renewcommand{\arraystretch}{1.15}
\centering
\fontsize{8pt}{10pt}\selectfont
\begin{tabular}{c|ccc|cc}
\multicolumn{1}{c}{} & \multicolumn{3}{c|}{Pseudo-healthy reconstruction (test\textsubscript{CN})} & \multicolumn{2}{c}{Anomaly detection (test\textsubscript{sAD})} \\ \hline
\multicolumn{1}{c}{} & MSE $(10^{-3})$ $\downarrow$ & PSNR $\uparrow$ & SSIM $\uparrow$ & IoU $\uparrow$ &  AP $\uparrow$ \\ \hline
\multicolumn{1}{c}{$\beta$-VAE} \cite{higgins_beta-vae_2017} &$17.02 \pm 3.21*$ & $23.84 \pm 0.83*$ & $69.98 \pm 3.57*$ & $22.82 \pm 5.31*$ & $28.05 \pm 8.43*$      \\ 
\multicolumn{1}{c}{f-AnoGAN} \cite{schlegl_f-anogan_2019} &  $31.70 \pm 8.87*$ & $21.23 \pm 1.11*$ & $59.78 \pm 5.63*$ & $17.56 \pm 4.65*$ & $21.36 \pm 7.82*$      \\ 
\multicolumn{1}{c}{\makecell{AnoDDPM \\ {\scriptsize (simplex)}} \cite{wyatt_anoddpm_2022}} & $\bm{8.72} \pm 1.33$ & $\bm{26.81} \pm 0.60$ & $79.79 \pm 2.75*$ & $15.95 \pm 4.77*$ & $26.41 \pm 7.21*$   \\ 
\hline
\multicolumn{1}{c}{BFN} \cite{graves_bayesian_2024} & $38.48 \pm 4.43*$ & $20.30 \pm 0.47*$ & $54.38 \pm 2.69*$ & $8.46 \pm 1.68*$ & $11.27 \pm 2.29*$    \\ 
\multicolumn{1}{c}{AnoBFN w/o [C2]} & $23.53 \pm 3.39*$ & $22.45 \pm 0.59*$ & $66.07 \pm 3.68*$ & $19.99 \pm 3.82*$ & $28.95 \pm 6.88*$    \\  
\multicolumn{1}{c}{AnoBFN} (Ours) & $9.67 \pm 1.66$ & $26.28 \pm 0.72$ & $\bm{80.92} \pm 2.62$ & $\bm{31.79} \pm 7.74$ & $\bm{48.17} \pm 11.31$   \\

\hline
\end{tabular}
\label{tab:results}
\end{table}

\begin{figure}[t!]
    \centering
    \includegraphics[width=0.96\textwidth, clip, trim={0 8 0 0}]{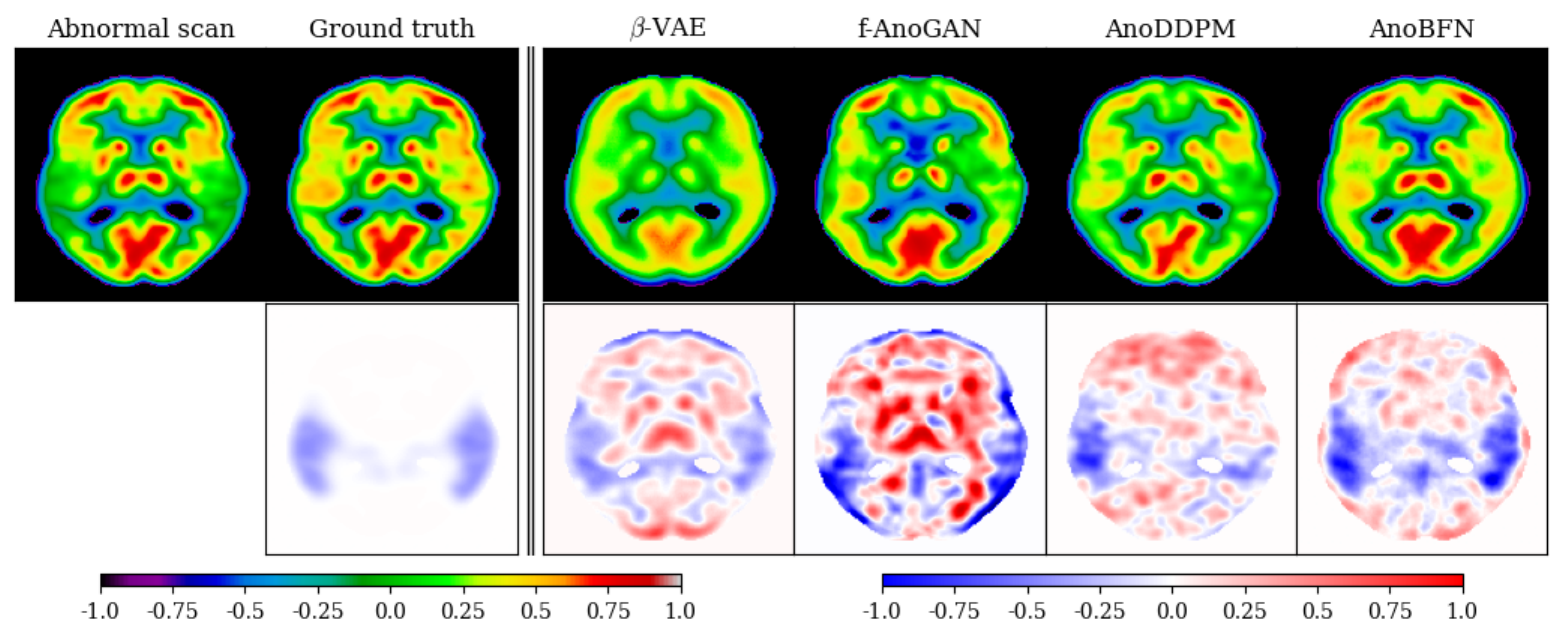}
    \caption{Example of reconstructions and residual maps for a random subject in the test set. Top: synthetic abnormal scan, ground truth (real scan without anomalies) and pseudo-healthy reconstructions generated by the different models from the synthetic abnormal scan. Bottom row: mask used to simulate hypometabolism and difference maps computed between each reconstruction and the abnormal scan.}
    \label{fig:visualization}
\end{figure}

\textit{Results.} Before evaluating performance on the primary task of anomaly detection, we first assess reconstruction performance on healthy data as a form of sanity check. While achieving high-quality reconstructions is not the primary objective, this step helps ensure that the models have learned meaningful representations of normal anatomy. As shown in Table~\ref{tab:results}, we can observe that both f-AnoGAN and $\beta$-VAE poorly reconstruct healthy images and underperform compared to AnoDDPM and AnoBFN, both producing similar results. AnoDDPM and f-AnoGAN are the methods the least able to detect anomalies, followed by $\beta$-VAE and AnoBFN. AnoBFN outperforms all models in AP and IoU ($p<0.01$). We also conducted an ablation study to assess each of our contributions to the original BFN. We can observe that introducing simplex noise and accuracy schedule [C1] (AnoBFN w/o [C2] vs BFN) leads to significant improvements in both healthy reconstruction and anomaly detection. Adding the Bayesian update guidance [C2] (AnoBFN) further improves our model, leading to state-of-the-art performance. Qualitative results are shown in Fig.~\ref{fig:visualization}. The $\beta$-VAE produces a blurred image, struggling to preserve high-uptake regions and fine details. f-AnoGAN, while not producing an exact reconstruction, effectively preserves the overall shape and shows higher contrast between normal and anomalous regions. AnoDDPM closely matches the original abnormal scan, preserves its structure, and captures a substantial portion of the anomaly. Finally, AnoBFN produces a sharp reconstruction that preserves subject's specificity while restoring healthy-appearing tissue in originally abnormal regions, leading to a clear delineation of anomalies. 

\section{Conclusion and future works}
We introduce Bayesian flow networks \cite{graves_bayesian_2024} to medical imaging for the first time, demonstrating their ability to perform unsupervised anomaly detection. To achieve this, we extended the original BFN framework to enable effective pseudo-healthy reconstruction and preservation of the subjects' identity.
AnoBFN introduces two key contributions: i) the combination of simplex noise and an accuracy schedule enabling conditional generation under high levels of spatially correlated noise, ii) a novel Bayesian update mechanism that recursively incorporates input information during inference, preserving subject specificity. Our approach outperforms state-of-the-art methods \cite{higgins_beta-vae_2017,schlegl_f-anogan_2019,wolleb_diffusion_2022} in the detection of synthetic anomalies simulating Alzheimer’s disease in FDG PET imaging. We confirmed the key role of the proposed Bayesian update by performing an ablation study. 
To enhance the robustness of anomaly detection, we aim to incorporate uncertainty quantification into the Bayesian generative process while also exploring alternative accuracy schedules and scaling metrics for the Bayesian update.
Beyond methodological refinements, we aim to evaluate the robustness and generalization of our approach by applying it to a broader range of medical imaging datasets, including diverse pathologies and imaging modalities.

\section{Acknowledgement}

The research leading to these results has received funding from the French government under management of Agence Nationale de la Recherche as part of the ``Investissements d'avenir'' program (references ANR-19-P3IA-0001 - PRAIRIE - and reference ANR-10-IAIHU-06) and the ``France 2030'' program (reference ANR-23-IACL-0008, PRAIRIE-PSAI), and from the European Union’s Horizon Europe Framework Program (grant number 101136607, project CLARA). This research is part of the MediTwin project, which has been funded by the French government as part of ``France 2030'' and by the European Union - Next Generation EU as part of the ``France Relance'' plan. R.D. received a Marie Skłodowska-Curie grant No 101154248 (project: SafeREG).

\bibliographystyle{splncs04}
\bibliography{Paper-0047}

\begin{thebibliography}{10}
\providecommand{\url}[1]{\texttt{#1}}
\providecommand{\urlprefix}{URL }
\providecommand{\doi}[1]{https://doi.org/#1}

\bibitem{atkinson_protein_2024}
Atkinson, T., Barrett, T.D., Cameron, S., Guloglu, B., Greenig, M., Robinson, L., Graves, A., Copoiu, L., Laterre, A.: Protein {Sequence} {Modelling} with {Bayesian} {Flow} {Networks} (2024). \doi{10.1101/2024.09.24.614734}

\bibitem{Baur2021AutoencodersUnsupervised}
Baur, C., Denner, S., Wiestler, B., Navab, N., Albarqouni, S.: Autoencoders for unsupervised anomaly segmentation in brain {{MR}} images: {{A}} comparative study. Medical Image Analysis p. 101952 (2021). \doi{10.1016/j.media.2020.101952}

\bibitem{cai2024rethinking}
Cai, Y., Chen, H., Cheng, K.T.: Rethinking autoencoders for medical anomaly detection from a theoretical perspective. In: Medical Image Computing and Computer-Assisted Intervention. pp. 544--554 (2024). \doi{10.1007/978-3-031-72120-5_51}

\bibitem{Cai2025MedIAnomalyComparative}
Cai, Y., Zhang, W., Chen, H., Cheng, K.T.: {{MedIAnomaly}}: {{A}} comparative study of anomaly detection in medical images. Medical Image Analysis  \textbf{102},  103500 (2025). \doi{10.1016/j.media.2025.103500}

\bibitem{graves_bayesian_2024}
Graves, A., Srivastava, R.K., Atkinson, T., Gomez, F.: Bayesian {Flow} {Networks} (2024), \url{http://arxiv.org/abs/2308.07037}

\bibitem{Hassanaly2024EvaluationPseudohealthy}
Hassanaly, R., Brianceau, C., Solal, M., Colliot, O., Burgos, N.: Evaluation of pseudo-healthy image reconstruction for anomaly detection with deep generative models: {{Application}} to brain {{FDG PET}}. Machine Learning for Biomedical Imaging  \textbf{2}(Special Issue for Generative Models),  611--656 (2024). \doi{10.59275/j.melba.2024-b87a}

\bibitem{Hassanaly2024PseudohealthyImage}
Hassanaly, R., Solal, M., Colliot, O., Burgos, N.: Pseudo-healthy image reconstruction with variational autoencoders for anomaly detection: {{A}} benchmark on {{3D}} brain {{FDG PET}} (2024), \url{https://inria.hal.science/hal-04445378}

\bibitem{higgins_beta-vae_2017}
Higgins, I., Matthey, L., Pal, A., Burgess, C.P., Glorot, X., Botvinick, M., Mohamed, S., Lerchner, A.: beta-{VAE}: {Learning} {Basic} {Visual} {Concepts} with a {Constrained} {Variational} {Framework}. In: International Conference on Learning Representations (2017), \url{https://openreview.net/forum?id=Sy2fzU9gl}

\bibitem{ho_denoising_2020}
Ho, J., Jain, A., Abbeel, P.: Denoising diffusion probabilistic models. In: Advances in Neural Information Processing Systems. vol.~33, pp. 6840--6851 (2020), \url{https://proceedings.neurips.cc/paper_files/paper/2020/file/4c5bcfec8584af0d967f1ab10179ca4b-Paper.pdf}

\bibitem{kingma_variational_2021}
Kingma, D., Salimans, T., Poole, B., Ho, J.: Variational diffusion models. In: Advances in Neural Information Processing Systems. vol.~34, pp. 21696--21707 (2021), \url{https://proceedings.neurips.cc/paper_files/paper/2021/file/b578f2a52a0229873fefc2a4b06377fa-Paper.pdf}

\bibitem{kingma_auto-encoding_2022}
Kingma, D.P., Welling, M.: Auto-{Encoding} {Variational} {Bayes} (2022). \doi{10.48550/arXiv.1312.6114}

\bibitem{loshchilov_decoupled_2019}
Loshchilov, I., Hutter, F.: Decoupled weight decay regularization. In: International Conference on Learning Representations (2019), \url{https://openreview.net/forum?id=Bkg6RiCqY7}

\bibitem{nichol_improved_2021}
Nichol, A., Dhariwal, P.: Improved {Denoising} {Diffusion} {Probabilistic} {Models} (Feb 2021). \doi{10.48550/arXiv.2102.09672}, \url{http://arxiv.org/abs/2102.09672}, arXiv:2102.09672 [cs]

\bibitem{pinaya_fast_2022}
Pinaya, W.H., Graham, M.S., Gray, R., Da~Costa, P.F., Tudosiu, P.D., Wright, P., Mah, Y.H., MacKinnon, A.D., Teo, J.T., Jager, R., et~al.: Fast unsupervised brain anomaly detection and segmentation with diffusion models. In: Medical Image Computing and Computer-Assisted Intervention. pp. 705--714 (2022). \doi{10.1007/978-3-031-16452-1_67}

\bibitem{ronneberger_u-net_2015}
Ronneberger, O., Fischer, P., Brox, T.: U-net: Convolutional networks for biomedical image segmentation. In: Medical Image Computing and Computer-Assisted Intervention (2015). \doi{/10.1007/978-3-319-24574-4_28}

\bibitem{routier_clinica_2021}
Routier, A., Burgos, N., D{\'i}az, M., Bacci, M., Bottani, S., {El-Rifai}, O., Fontanella, S., Gori, P., Guillon, J., Guyot, A., Hassanaly, R., Jacquemont, T., Lu, P., Marcoux, A., Moreau, T., {Samper-Gonz{\'a}lez}, J., Teichmann, M., {Thibeau-Sutre}, E., Vaillant, G., Wen, J., Wild, A., Habert, M.O., Durrleman, S., Colliot, O.: Clinica: {{An Open-Source Software Platform}} for {{Reproducible Clinical Neuroscience Studies}}. Frontiers in Neuroinformatics  \textbf{15}, ~39 (2021). \doi{10.3389/fninf.2021.689675}

\bibitem{ruple2025symmetryawarebayesianflownetworks}
Ruple, L., Torresi, L., Schopmans, H., Friederich, P.: Symmetry-aware bayesian flow networks for crystal generation (2025), \url{https://arxiv.org/abs/2502.03146}

\bibitem{schlegl_f-anogan_2019}
Schlegl, T., Seeböck, P., Waldstein, S.M., Langs, G., Schmidt-Erfurth, U.: f-{AnoGAN}: {Fast} unsupervised anomaly detection with generative adversarial networks. Medical Image Analysis  \textbf{54},  30--44 (2019). \doi{10.1016/j.media.2019.01.010}

\bibitem{song_score-based_2021}
Song, Y., Sohl-Dickstein, J., Kingma, D.P., Kumar, A., Ermon, S., Poole, B.: Score-based generative modeling through stochastic differential equations. In: International Conference on Learning Representations (2021), \url{https://openreview.net/forum?id=PxTIG12RRHS}

\bibitem{thibeau-sutre_clinicadl_2022}
Thibeau-Sutre, E., Díaz, M., Hassanaly, R., Routier, A., Dormont, D., Colliot, O., Burgos, N.: {ClinicaDL}: {An} open-source deep learning software for reproducible neuroimaging processing. Computer Methods and Programs in Biomedicine  \textbf{220},  106818 (2022). \doi{10.1016/j.cmpb.2022.106818}

\bibitem{weiner_alzheimers_2010}
Weiner, M.W., Aisen, P.S., Jack, C.R., Jagust, W.J., Trojanowski, J.Q., Shaw, L., Saykin, A.J., Morris, J.C., Cairns, N., Beckett, L.A., Toga, A., Green, R., Walter, S., Soares, H., Snyder, P., Siemers, E., Potter, W., Cole, P.E., Schmidt, M.: The {Alzheimer}'s {Disease} {Neuroimaging} {Initiative}: {Progress} report and future plans. Alzheimer's \& Dementia  \textbf{6}(3), ~202 (2010)

\bibitem{weiner_alzheimers_2017}
Weiner, M.W., Veitch, D.P., Aisen, P.S., Beckett, L.A., Cairns, N.J., Green, R.C., Harvey, D., Jack, C.R., Jagust, W., Morris, J.C., Petersen, R.C., Salazar, J., Saykin, A.J., Shaw, L.M., Toga, A.W., Trojanowski, J.Q.: The {Alzheimer}'s {Disease} {Neuroimaging} {Initiative} 3: {Continued} innovation for clinical trial improvement. Alzheimer's \& Dementia  \textbf{13}(5),  561--571 (2017)

\bibitem{wolleb_diffusion_2022}
Wolleb, J., Bieder, F., Sandk{\"u}hler, R., Cattin, P.C.: Diffusion models for medical anomaly detection. In: Medical Image Computing and Computer-Assisted Intervention. pp. 35--45 (2022). \doi{10.1007/978-3-031-16452-1_4}

\bibitem{wu2025periodicbayesianflowmaterial}
Wu, H., Song, Y., Gong, J., Cao, Z., Ouyang, Y., Zhang, J., Zhou, H., Ma, W.Y., Liu, J.: A periodic bayesian flow for material generation (2025), \url{https://arxiv.org/abs/2502.02016}

\bibitem{wyatt_anoddpm_2022}
Wyatt, J., Leach, A., Schmon, S.M., Willcocks, C.G.: {AnoDDPM}: {Anomaly} {Detection} with {Denoising} {Diffusion} {Probabilistic} {Models} using {Simplex} {Noise}. In: 2022 {IEEE}/{CVF} {Conference} on {Computer} {Vision} and {Pattern} {Recognition} {Workshops}. pp. 649--655 (2022). \doi{10.1109/CVPRW56347.2022.00080}

\bibitem{yu_adversarial_2023}
Yu, J., Oh, H., Yang, J.: Adversarial denoising diffusion model for unsupervised anomaly detection. In: Deep Generative Models for Health Workshop NeurIPS 2023 (2023), \url{https://openreview.net/forum?id=PvHhhn1iX9}

\end{thebibliography}

\end{document}